\pdfoutput=1
\documentclass[11pt]{article}
\usepackage{amsfonts}

\usepackage{amsmath}
\usepackage{booktabs}
\usepackage{tabu}
\usepackage[T1]{fontenc}
\usepackage{colortbl}
\usepackage{graphicx}
\usepackage{booktabs}
\usepackage[table,xcdraw]{xcolor}
\usepackage[normalem]{ulem}
\usepackage{subcaption}
\useunder{\uline}{\ul}{}
\usepackage[final]{acl}

\usepackage{times}
\usepackage{latexsym}

\usepackage[T1]{fontenc}
\usepackage[utf8]{inputenc}

\usepackage{microtype}

\usepackage{inconsolata}

\usepackage{graphicx}

\title{Mitigating Memorization in LLMs using Activation Steering}

\author{Manan Suri, Nishit Anand, Amisha Bhaskar 
\\ University of Maryland, College Park
\\ \texttt{\{\href{mailto:manans@umd.edu}{manans}, \href{mailto:nishit@umd.edu}{nishit}, \href{mailto:amishab@umd.edu}{amishab}\} @umd.edu}  }

\begin{document}
\maketitle
\begin{abstract}
The memorization of training data by Large Language Models (LLMs) poses significant risks, including privacy leaks and the regurgitation of copyrighted content. Activation steering, a technique that directly intervenes in model activations, has emerged as a promising approach for manipulating LLMs. In this work, we explore the effectiveness of activation steering in reducing memorization while preserving generalization capabilities. We conduct empirical evaluations using a controlled memorization benchmark of literary material and demonstrate that our method successfully suppresses memorized content with minimal degradation in model performance in Gemma. Additionally, we analyze the trade-offs between suppression effectiveness and linguistic fluency, highlighting the advantages and limitations of activation-based interventions. Our findings contribute to ongoing efforts in developing safer and more privacy-preserving LLMs by providing a practical and efficient mechanism to mitigate unintended memorization.
\end{abstract}

\section{Introduction} 
% \todo[color=pink]{Manan}\todo[]{The third paragraph is confusing to me: “low semantic or behavioral impact” are not explained or defined. Could you properly define the terms, and also give some intuition there?}  \todo[color=green]{Lets not mention semantic features here, we can mention that in a later chapter}
Large Language Models (LLMs) have become increasingly effective at generating coherent and contextually relevant text, but they face a critical issue: the tendency to memorize specific training data, rather than generalize from it. Memorization in LLMs poses a number of problems, including potential privacy violations, biased output generation, and reduced adaptability to novel contexts. When LLMs memorize exact sequences or phrases from training data, they may inadvertently reveal sensitive information or reproduce biases embedded in the data, undermining ethical and practical objectives in deploying these models. \label{ch_1}

Recent studies have shown, that memorized sequences are intrinsically tied to the model's weights \cite{chang2024localization, nasr2023scalable,stoehr2024localizing}, and several studies have been successful at eliminating specific neurons that lead to memorization of specific sequences \cite{chang2024localization,stoehr2024localizing,ippolito2022preventing}. Based on these intuitions, we want to explore generalized mitigation of memorization in LLMs using manipulation of the model's weighted activations in the forward pass. However, directly altering model weights is impractical and risks destabilizing model performance. Activation steering provides a potential solution: guiding selected activations rather than arbitrarily altering the weights allows for controlled mitigation of memorized content without compromising the model’s overall abilities.

Our hypothesis is  that activation steering can help reduce memorization without impairing the model's overall abilities under certain conditions. In this study, we aim to analyze if activation steering can be used as a practical method to mitigate memorization in LLMs, and study the exact factors that influence this. Specifically, we aim to answer the following research questions:

\textbf{RQ1:} Can activation steering mitigate memorization in LLMs? Is this a general behaviour, or do specific features matter?

\textbf{RQ2:} What is the impact of different factors in our activation steering-based system on memorization, linguistic ability, and general abilities?

\textbf{RQ3:} What are best practices while using activation steering for memorization mitigation that avoid undesired outcomes of the method? 

For \textbf{RQ1} and \textbf{RQ2}, we perform quantitative experiments with different features, in varying configurations (strength, model layer) on different benchmarks. For \textbf{RQ3} we take the findings from \textbf{RQ1} and \textbf{RQ2}, along with insights from qualitative examples (such as finding and defining neurons with high semantic footprint) to assemble a guide for using activation steering to mitigate model memorization in practical settings.
% targets features with low semantic or behavioral impact, thereby minimizing unintended behavioral shifts. Steering high-semantic features can lead to specific, sometimes unpredictable behaviors, such as adopting stylized modes of output.  Avoiding such high-semantic features during activation steering, while selecting low-impact features, may prevent these unwanted outputs, balancing the trade-off between mitigation of memorized sequences and model performance. 

\section{Related Work}
\subsection{Influences and Dynamics of Memorization}

Memorization in large language models (LLMs) has been a subject of considerable research, with several studies uncovering the influences and dynamics of this phenomenon. Satvaty et al. \cite{satvaty2024undesirable} provide a detailed survey emphasizing the impact of model characteristics such as size and the specific architectures on memorization. They note that these factors significantly affect how memorization manifests in different LLMs, underlining the importance of architectural features in managing this phenomenon (Satvaty et al., 2024 \cite{satvaty2024undesirable}). Carlini et al. (2021, 2023) \cite{carlini2021extracting,carlini2022quantifying} and Lee et al. (2023) \cite{lee2021deduplicating} have noted that LLMs, such as GPT models, can verbatim memorize sequences from their training corpora, leading to potential privacy and copyright issues. In addition, Nasr et al. (2023) \cite{nasr2023scalable} highlight that model capacity and the frequency of data during training contribute significantly to verbatim memorization, emphasizing the need for strategies that mitigate these risks without sacrificing model utility.

Leybzon and Kervadec \cite{leybzonlearning} explore the temporal dynamics of memorization, observing that memorization rates vary throughout the training process, with peaks at the beginning and end. This observation suggests that memorization is influenced by the training dynamics, where data may undergo cycles of being memorized, forgotten, and then re-memorized, depending on how it is presented during the training sequence.

A novel approach to understanding memorization involves the localization of memorized content within LLMs. Chang et al. (2024) \cite{chang2024large} introduce methods for identifying specific model weights and neurons responsible for memorization. Their work suggests that localization can be a crucial step towards "neural surgery"—the selective editing or removal of model components to eliminate memorized data, thus enhancing privacy and model safety without broadly impacting model performance. This approach is particularly relevant as it diverges from the notion that memorization is distributed non-specifically across model parameters.

Recent benchmarks by Chang et al. (2024) \cite{chang2024localization} offer a systematic evaluation of localization techniques, illustrating that methods derived from network pruning, such as HARD CONCRETE, show promise in effectively identifying critical neurons involved in memorization. These findings indicate that while certain neurons are integral to memorizing specific sequences, they also play roles in general language tasks, which complicates the task of memorization removal.

Moreover, the dual benchmarks introduced—INJ for ground-truth localization and DEL for practical application—help elucidate the effectiveness of different localization methods under controlled and real-world conditions. The results from these benchmarks demonstrate that while localization can pinpoint the neurons contributing to memorization, removing these neurons often affects the memorization of other sequences as well, underscoring the intertwined nature of neural functions within LLMs. An overview of the findings for each
of the factors discussed in this subsections is provided in
Table I.

% Focusing on the role of model components, Huang et al. challenge the notion that specific weights or neurons are solely responsible for verbatim memorization. Their studies suggest that verbatim memorization results from complex interactions within distributed model states rather than isolated components. They argue that these memorizations leverage the general capabilities of language models, indicating that mitigating verbatim memorization without impacting overall performance might be challenging (Huang et al., 2024).

% The role of localization in memorization, specifically how particular neurons and weights contribute to this process, is crucial for developing targeted mitigation strategies. Research by Chang et al. (2024) and Stoehr et al. (2024) attempts to localize the memorization to specific components like attention heads or individual neurons, proposing that interventions targeted at these localized features could potentially erase unwanted memorizations while preserving the utility of the model. However, the efficacy of these interventions remains limited, suggesting a more integrated approach may be necessary to address the underlying complexities of memorization in LLMs effectively.

\begin{table*}[ht]
\label{tab:influence}
\centering

\label{tab:memorization_influences}
\resizebox{0.8\textwidth}{!}{%
\begin{tabular}{|>{\raggedright\arraybackslash}m{4cm}|>{\raggedright\arraybackslash}m{7cm}|>{\raggedright\arraybackslash}m{4cm}|}
\hline
\textbf{Factor from Section IV \cite{satvaty2024undesirable}} & \textbf{Key Findings} & \textbf{References} \\ \hline
Model capacity & Larger models memorize more & Carlini et al. \cite{carlini2022quantifying}, Tirumala et al. \cite{tirumala2022memorization} \\ \hline
Training data characteristics & Duplicated data amplifies memorization & Kandpal et al. \cite{kandpal2022deduplicating}, Lee et al. \cite{lee2021deduplicating} \\ \hline
Input and prompting strategies & Longer prompts and prompt tuning can facilitate recall of the memorized suffix. & Carlini et al. \cite{carlini2021extracting}, McCoy et al. \cite{mccoy2023much}, Ozdayi et al. \cite{ozdayi2023controlling} \\ \hline
Tokenization & Bigger tokenizer vocabulary leads to more memorization & Kharitonov et al. \cite{kharitonov2021bpe} \\ \hline
Sampling methods & While greedy sampling can pinpoint extremely memorized samples, top-n sampling is the most effective method to retrieve more memorized items. & Carlini et al. \cite{carlini2021extracting}, Yu et al. \cite{yu2023bag} \\ \hline
Fine-tuning & The amount of memorization after fine-tuning significantly varies depending on the task. & Zeng et al. \cite{zeng2023exploring}, Mireshghallah et al. \cite{mireshghallah2022quantifying} \\ \hline
Training process dynamics & Earlier phases of training are less prone to memorization & Kandpal et al. \cite{kandpal2022deduplicating}, Zhang et al. \cite{zhang2023counterfactual}, Jagielski et al. \cite{jagielski2022measuring} \\ \hline
Forgetting mechanisms & Forgetting follows an exponentially decaying curve & Tirumala et al. \cite{tirumala2022memorization}, Jagielski et al. \cite{jagielski2022measuring} \\ \hline
\end{tabular}%
}

\caption{Influences and Dynamics of Memorization and Their Key Findings \cite{satvaty2024undesirable}}
\end{table*}

\subsection{Mitigation Techniques for Memorization in LLMs}

In the domain of large language models (LLMs), memory mitigation is crucial for addressing privacy concerns and avoiding inadvertent data regurgitation. Several innovative approaches have been proposed to tackle this issue. The "Goldfish Loss" \cite{hans2024like} method introduces a training modification that prevents the model from learning random subsets of tokens, effectively reducing exact data memorization. Similarly, the MemoAnalyzer tool \cite{zhang2024ghost} enhances privacy by allowing users to manage and modify sensitive data identified during interactions. The Self-Synthesized Rehearsal (SSR) \cite{huang2024mitigating} technique uses synthetic data generation for model training, which preserves performance without compromising privacy. The Parameterized User Memory Injection (MiLP) \cite{zhang2024personalized} approach maintains user personalization by integrating historical content into a learnable representation within LLMs. 
Furthermore, the Bayesian simulator MemSim \cite{zhang2024memsim} generates reliable evaluation datasets to assess the memory capabilities of personal assistant models without using real user data.

Expanding on the theme of privacy, Differentially Private (DP) \cite{abadi2016deep} training offers robust protections by minimizing the impact of any single data point on the model’s output, albeit at the cost of potential reductions in model utility and increased resource requirements. Techniques like pretraining on sanitized data before DP training can enhance the feasibility of these methods. Additional strategies include data deduplication \cite{kandpal2022deduplicating} and the use of Bloom filters \cite{ippolito2022preventing,anil2021large} for detecting memorization at test time, although they face challenges due to the complexity of web data and the risk of missing duplicates. The research on LLMs as knowledge bases and memorization of specialized content, such as efficient algorithms or writing styles, suggests a broader perspective on memorization beyond verbatim recall. Emerging areas like distribution inference and the alignment of LLMs to specific goals also contribute to a deeper understanding of how LLMs manage and utilize learned information, although comprehensive studies explicitly focusing on memorization are still scarce. 

\subsection{Activation Steering}
\cite{turner2023activation} introduced the Activation Addition approach, which generates steering vectors by computing the difference in activations between a pair of prompts at a specific layer and token position in a transformer model. These steering vectors are then applied to influence the model’s completions by modifying the first token position in forward passes. However, this approach has limitations, including inconsistent performance across different prompts and behaviors, limited robustness, and an evaluation restricted to GPT-2-XL. By contrast, methods utilizing a dataset of diverse contrast pairs instead of a single pair can enhance the precision and reliability of steering vector encodings, allowing for a more robust control over model behavior across varied prompts.

\cite{li2024inference} use linear probes on a contrastive question-answering dataset to predict truthfulness, identifying attention heads associated with truthful responses. They utilize a Mean Difference vector between true and false distributions to shift activations, which improves truthfulness with minimal impact on fluency. This technique requires relatively little data and is effective on adversarial benchmarks. \cite{zou2023representation} further explore techniques for extracting representations of high-level concepts, such as honesty and emotion, in large language models (LLMs). \cite{panickssery2023steering} scaled the method by performed contrastive steering on LLaMa2. \cite{lee2021deduplicating} introduces Conditional Activation Steering (CAST), a technique that examines activation patterns in large language models (LLMs) during inference to selectively apply or withhold activation steering based on the context of the input. Recently, \cite{templeton2024scaling,chalnev2024improvingsteeringvectorstargeting} have introduced methods which use interpretable Sparse Autoencoder (SAE) based features for activation steering. 

\cite{chang2024localization} focuses on identifying the exact weights and neurons responsible for memorization using specially designed benchmarks. It evaluates how effectively various localization methods can pinpoint these specific model components that store memorized information. On the other hand our method aims to mitigate memorization by manipulating the model's activations during the forward pass to steer clear of high memorization areas, without directly identifying specific weights or neurons involved in memorization. \cite{ippolito2022preventing} uses a direct intervention approach during output generation, where it filters out memorized content using a Bloom filter to prevent the generation of memorized sequences. While, we focus on a subtler manipulation of activation patterns to prevent the model from accessing memorized information without the need for output filtering or explicit blocking of content generation. \cite{stoehr2024localizing} Employs a detailed analysis of model parameters and gradients to pinpoint the exact sources of memorization within the model's structure, focusing on editing and altering these memorized contents. Whereas our method, rather than identifying specific components or altering them, it aims to steer the model's activation patterns during normal operation, influencing how the model processes information to reduce memorization risk without changing the model's structure or specific parameters.

% \todo[color=pink]{Amisha}
% \todo{Related Work: what is the difference between the proposed activation steering and [8, 13, 28]?}
% \todo[color=green]{Mention the difference between present memory-based memorization mitigation techniques and activation steering.}

\section{Method}
\subsection{Background}
\textbf{Feature Steering using SAE Derived Interpretable Features}

A sparse autoencoder (SAE) \cite{ng2011sparse} is a neural network architecture designed to learn a compressed representation of input data, with a sparsity constraint applied to the hidden layer activations to encourage the discovery of distinct and interpretable latent features. SAEs have shown great potential as a tool to extract sparse and disentangled representations of high-dimensional model weights, to identify interpretable features within the model’s hidden layer activations. SAEs constrain the activations to be sparsely distributed, aligning with the hypothesis that many natural latent variables in models are sparse. This sparsity enables the SAE to capture specific, low-dimensional features embedded in a high-dimensional space, allowing for the decomposition of the model’s hidden layer activations into distinct, interpretable components. Each component can be viewed as an abstraction that captures certain attributes of the input data, facilitating a form of feature disentanglement and enabling more granular analysis and control of the model's internal representations. \cite{cunningham2023sparse}

We build upon recent work which leverages SAE features for activation steering \cite{templeton2024scaling, chalnev2024improvingsteeringvectorstargeting}. Let \( \mathbf{a} \in \mathbb{R}^d \) represent the activations at a particular layer in the model, where \( d \) is the dimensionality of the activation space. After learning a sparse representation, the sparse autoencoder (SAE) provides a set of feature vectors, which can be derived for each index of the SAE, with an SAE corresponding to each MLP layer in the LLM, by simply decoding the vector corresponding to that index in that layer's SAE.

The steering process adjusts the activations of a layer, by adding a scaled version of the steering vectors derived from the SAE features derived from a particular index of the SAE of that layer. The steering operation can be mathematically formulated as follows:

\begin{equation}
   \mathbf{a}_{\text{steered}} = \mathbf{a} + \alpha \cdot \beta \cdot \mathbf{v}_i 
\end{equation}

where:
\begin{itemize}
\item  \( \mathbf{a} \in \mathbb{R}^d \) is the original activation vector,
\item  \( \mathbf{a}_{\text{steered}} \in \mathbb{R}^d \) is the modified activation vector after steering,
\item  \( \alpha \in \mathbb{R} \) is a scaling factor, given by the maximum observed activation value that normalizes the effect size of the steering,
\item  \( \beta \in \mathbb{R} \) is a steering strength coefficient controlling the influence of the feature on the activations,
\item  \( \mathbf{v}_i \in \mathbb{R}^d \) is the steering vector associated with the \( i \)-th index of the SAE vector, which is decoded from the sparse autoencoder.
\end{itemize}

\textbf{Auto-interpretation of LLM Features}
Auto-interpretation involves leveraging advanced language models like GPT-4 to identify the real-world features represented by the learned features of a Sparse Autoencoder (SAE). This process is achieved by visualizing the activation maps of the SAE features on various text samples, allowing for an exploration of how specific features correlate with different linguistic patterns or concepts. Once the activations are mapped, the model is prompted to generate possible explanations for each feature, thereby providing a semantic interpretation of the abstract activations. This approach offers a way to automatically assign meaning to the latent variables, enhancing the interpretability of complex neural network representations. \cite{bills2023language}

\subsection{Methodology}

To evaluate the impact of activation steering on model memorization and performance, we conduct a comprehensive grid-style experiment with the following experimental axes:  

\begin{enumerate}
    \item \textbf{Layer of the Model:} The intervention is applied at various layers within the model to assess how steering effects differ across the model's hierarchy.  
    \item \textbf{Specific Feature:} Features are selected from the Sparse Autoencoder (SAE) representation, without any filtering or prioritization, to ensure the experiments cover a broad and unbiased feature space.  
    \item \textbf{Strength $\beta$ of Steering:} The magnitude of the activation steering intervention is varied to observe its influence on both memorization and performance metrics.  
\end{enumerate}

Random selection of features ensures an unbiased exploration of the model's internal representation space, avoiding overfitting experimental results to pre-identified features with known properties. This approach minimizes confirmation bias, promotes generalizability by preventing reliance on specific assumptions, and allows for broader coverage of the Sparse Autoencoder’s feature space, including less prominent dimensions that may influence model behavior. Additionally, automating feature selection reduces manual effort, enabling large-scale and systematic experiments.

A large-scale set of experiments covers combinations of the above parameters. To maintain the analysis's generalizability and robustness, no filtering or manual selection is performed.  

Each experiment involves a steering-based intervention on the model, followed by evaluations on:  
\begin{enumerate}
    \item \textbf{Memorization Benchmark:} To measure the reduction in the model's tendency to memorize training data.  
    \item \textbf{Linguistic Modeling Abilities:} To assess the model's ability to generate coherent and meaningful text.  
    \item \textbf{LLM Abilities:} To evaluate the overall performance on tasks requiring reasoning, understanding, and contextual awareness.  
\end{enumerate}

Additionally, control experiments are performed without steering interventions across all experimental configurations to serve as a baseline for comparison.

\section{Results}
\subsection{Data}
We curated a dataset of 40 books, their metadata, and their opening lines. These 40 books were sampled from the top 1000 novels on GoodReads, a book review platform. Since several of these books are protected by copyrights, we had to manually collect the opening sentences of the books.  Table \ref{tab:books_dataset} shows the list of books in our dataset.

% \todo[color=pink]{Manan}
% \todo{Why are the semantics and behavior of the features relevant here?}
% \todo[color=green]{New section: Practical Guide/ Practical Outcomes}

\begin{table*}[]
\centering
\renewcommand{\arraystretch}{2} 
\resizebox{\linewidth}{!}{%
\begin{tabular}{|l|l|l|l|}
\hline
Pride and Prejudice & Dracula & The Hound of the Baskervilles & Fahrenheit 451 \\ \hline
The Yellow Wallpaper & The Prince & Three Men in a Boat & Invisible Man \\ \hline
Alice's Adventures in Wonderland & The Picture of Dorian Gray & The Great Gatsby & The Stranger \\ \hline
Frankenstein; Or The Modern Prometheus & War and Peace & Love in the Time of Cholera & The Bell Jar \\ \hline
Metamorphosis & A Tale of Two Cities & Harry Potter and the Philosopher's Stone & To Kill a Mockingbird \\ \hline
Adventures of Huckleberry Finn & Les Misérables & 1984 & The Handmaid's Tale \\ \hline
The Importance of Being Earnest & The Jungle Book & Animal Farm & Norwegian Wood \\ \hline
The Adventures of Tom Sawyer & Crime and Punishment & One Hundred Years of Solitude & The Hobbit \\ \hline
Great Expectations & The Iliad & Lolita & The Lion, the Witch and the Wardrobe \\ \hline
Ulysses & The Wonderful Wizard of Oz & Moby Dick & Gulliver's Travels \\ \hline
\end{tabular}%
}
\caption{List of books included in the memorization dataset curated by us.}
\label{tab:books_dataset}
\end{table*}

\subsection{Experiments}

% \todo[color=pink]{Manan}
% \todo[color=green]{Experiment Space: Layers, features (indices), strength}
% We perform the following experiments:

\subsubsection{Memorization}
    
    This experiment tests the ability of the LLM to recall the first few sentences of famous novels, under control and treatment scenarios, i.e. with or without activation steering. These were evaluated on the dataset described above. \label{mem_expt}
    
    The model is prompted to return the first few lines of the book, and the response is compared to the ground truth lines. We use a simple prompt, given by:
\vspace{3pt}

        \fbox{
        \centering
        \begin{minipage}{0.8\linewidth}
        \centering
        \textit{Do you know the first few lines of {Book Name}?}
        \newline
        \textit{JUST RETURN THE FIRST FEW LINES. DO NOT ADD ADDITIONAL TEXT.} 
        \end{minipage}
        }

    \begin{enumerate}
        % \item \textbf{Word Overlap F1:} 
        % Word Overlap F1 refers to the average overlap between the model’s output and the ground truth sentence. F1-Score calculates precision and recall at word level and combines them in a single score.
        % \begin{equation}
        % \text{F1} = \frac{2 \cdot \text{precision} \cdot \text{recall}}{\text{precision} + \text{recall}}
        % \end{equation}
        
        % \begin{equation}
        % \text{precision} = \frac{\text{TP}}{\text{TP} + \text{FP}}
        % \end{equation}
        
        % \begin{equation}
        % \text{recall} = \frac{\text{TP}}{\text{TP} + \text{FN}}
        % \end{equation}
        % where,
        % True Positive (TP) is the number of words which overlap between the ground truth sentence and the model’s output.
        % False Positive (FP) is the number of words present in the model’s output which are missing in the ground truth. 
        % False Negative (FN) is the number of words present in the ground truth but missing in the model’s output.

        \item \textbf{ANLCS:} 
        ANLCS is Average Normalized Longest Common Subsequence. For every example, LCS is the longest subsequence which is present in both the ground truth sentence and the model’s output. We normalize it by dividing it by the length of the ground truth sentence, and then take the average of this value over all samples. This is how we calculate ANLCS.

    \end{enumerate}

% \todo[color=pink]{Nishit}
% \todo[color=green]{New Experiment: Language Modeling}

 \subsubsection{Language Modeling}
    
To test the impact of our system parameters in activation steering on the language modeling abilities of the LLM, we perform an experiment using the Microsoft Research Paraphrase Corpus (MRPC)\cite{dolan2005automatically}. This is motivated by the fact that sentence-level paraphrasing is a task that involves inferring semantic and lexical context of the input, and generating a similar sentence; therefore a linguistically incorrect sentence would tend to be inconsistent with the ground truth source and paraphrase sentences. The MRPC dataset consists of human-annotated sentence pairs extracted from NewsWire articles, indicating whether the pair is a paraphrase. We select 50 samples from this corpus for our experiments. \label{lm_expt}
The following prompt was used:

\vspace{3pt}
\fbox{
        \centering
        \begin{minipage}{0.8\linewidth}
        \centering
        \textit{You are given the following sentence. Paraphrase the sentence, keeping}
        \newline
        \textit{the meaning of the sentence same. <Test sentence>} 
        \end{minipage}
        }

\vspace{5pt}
 
    We compared the generated paraphrases from model with the ground truth sentences using BERTScore\cite{zhang2019bertscore} and METEOR\cite{banerjee2005meteor}. Additionally, we judge the perplexity of these models independently, as a stand-alone metric for language modeling ability. 
    \begin{enumerate}
        \item \textbf{BERTScore}
    
    BERTScore is used to assess the quality of text generation by measuring the semantic similarity between a candidate text, like the model's output and a reference text, like the ground truth. It utilizes contextual embeddings from pre-trained transformer models like BERT, which allows it to focus on meaning rather than surface-level word overlap.
    
    BERTScore offers an advantage over traditional metrics like BLEU and ROUGE by focusing on contextual meaning rather than relying solely on n-gram overlaps. This makes it useful for tasks like summarization, paraphrasing, or translation, where semantic similarity is crucial and variations in wording, like synonyms, are common.

        \item  \textbf{METEOR}

    METEOR (Metric for Evaluation of Translation with Explicit ORdering) is designed to overcome the limitations of traditional metrics like BLEU and ROUGE for the assessment of text generation. Originally designed for machine translation, it is widely used for tasks such as summarization and paraphrasing. METEOR evaluates the alignment between a candidate text, i.e., the model’s output, and a reference text, i.e., ground truth text, by incorporating linguistic features such as synonym matching, stemming, and word order consideration. This linguistically-informed approach yields evaluations that better align with human judgment.

    \item \textbf{Perplexity}
    
    Perplexity evaluates the performance of a language model by assessing how accurately it predicts a given sequence of text. It measures the model's level of uncertainty or "surprise" when processing the actual sequence of words. A lower perplexity score indicates that the model assigns higher probabilities to the correct words, reflecting stronger predictive performance. Perplexity is widely used to assess their ability to generate fluent and coherent text. We use GPT2 and its tokenizer to calculate perplexity. High perplexity means that the text distribution is "surprising", therefore, lower values are better by comparison.
    \end{enumerate}

   \subsubsection{General Performance}
    We test the LLM's performance on standard LLM benchmark tasks with and without activation steering, to measure the impact of activation steering on LLM performance. \label{perf_expt}
    \textbf{Benchmarks}
    The benchmakrs used have been described below. 
    % Samples from each of the perfomance evaluation sets have been represented in table \ref{tab:sample_questions}.
    \begin{enumerate}
        \item \textbf{BIG-Bench Hard (BBH)}
        BIG-Bench Hard \cite{suzgun-etal-2023-challenging} is a standard benchmark for measuring the performance of LLMs on various reasoning, arithmetic and linguistic tasks. 

        From BBH, we evaluate performance on the following tasks:

        \begin{itemize}
            \item \textbf{Boolean Expressions}: It contains questions which consist of multiple True and False Boolean Constants and multiple (or, and, not) boolean operators, and the model has to evaluate the truth value of the Boolean expression and give answer as True or False.
            \item \textbf{Date Understanding}: In this task, some sentences are given regarding a particular date in the question and the model has to calculate the correct date and select the correct option.
            \item \textbf{Logical Deduction}: In this task, the question gives information about the spatial relationship and placement of a few objects and the model has to deduce the order of sequence of those objects and select the correct option.
            \item \textbf{Snarks}: In this task, two almost identical sentences are given in the question and the model has to find which one out of these two is sarcastic.
        \end{itemize}

        \item \textbf{BoolQ}
        BoolQ \cite{clark2019boolqexploringsurprisingdifficulty} is an Natural Language Understanding (NLU) benchmark to measure the ability of Language Models (LMs) in Yes/No questions based on common-knowledge questions.

\end{enumerate}

\subsection{Experimental Set-up}

Our experiments are conducted using NeuronPedia \cite{neuronpedia}, a mechanistic interpretability platform designed to assist researchers working with Sparse Autoencoders (SAEs). NeuronPedia provides access to models, feature dashboards, data visualizations, and tools for conducting and analyzing experiments. Specifically, we utilize their Steering and Feature Search (Auto-Interpretability) API endpoints to execute our methodology.  

The experiments are performed on the \texttt{Gemma-2-9B-IT} model, with $n=100$ experiments conducted in a random grid configuration.  \label{new_expts}

The random parameters for each run are are defined as follows: 

\begin{itemize}
    \item \textbf{Layer:} Randomly selected from \{9, 20, 31\}.  
    \item \textbf{Strength:} Randomly sampled from the range [-100, 100].  
    \item \textbf{Feature:} For each layer, a feature index is randomly sampled from the range [0, 131072), representing the 131k features available per layer.  
\end{itemize}

The temperature is set at 0.5. For each set of parameters, a default, unsteered generation is also done.

\begin{figure*}[ht!]
    \centering
    % Subfigure 1
    \begin{subfigure}{0.50\textwidth}
        \includegraphics[width=\textwidth]{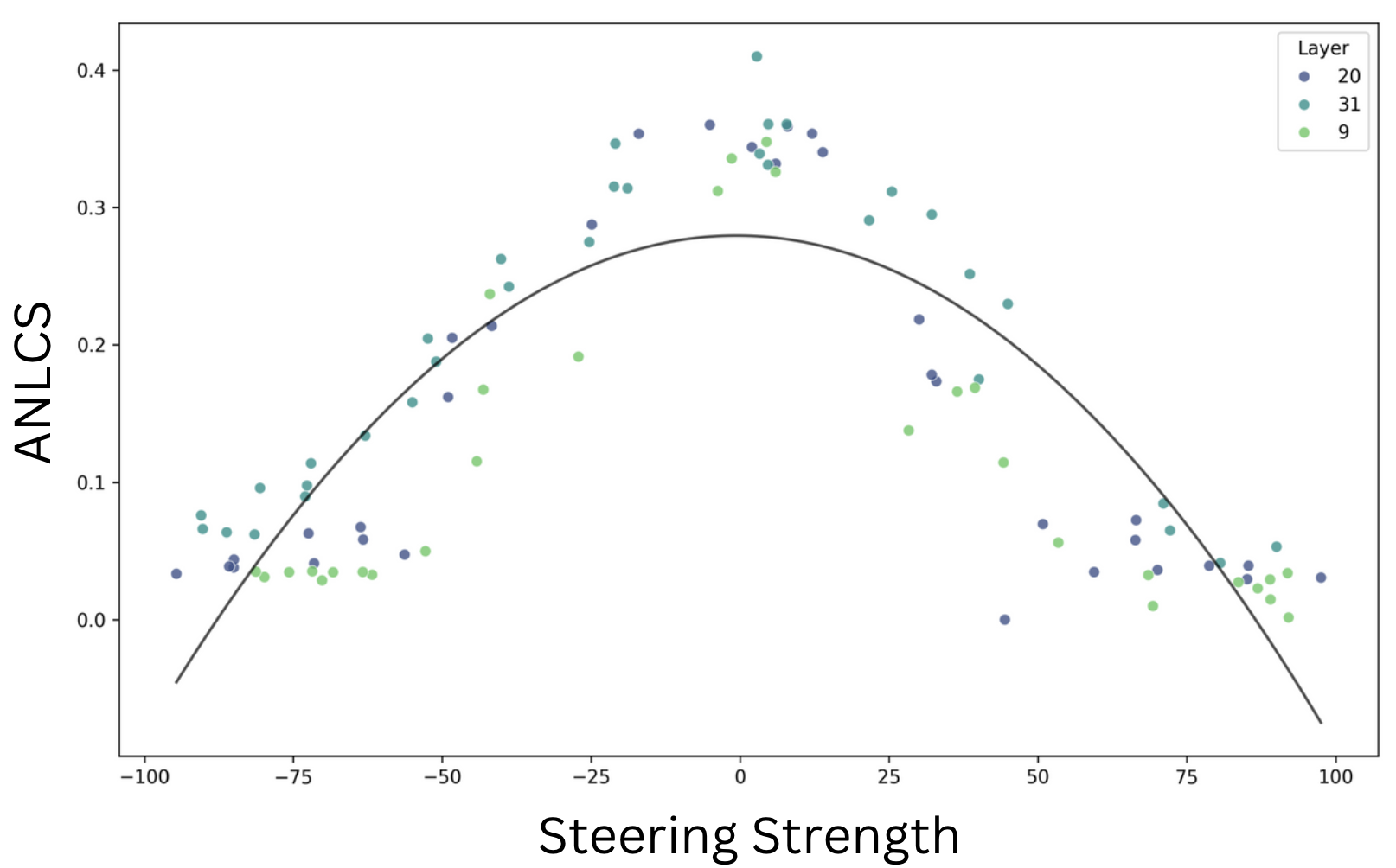}
        \caption{Steered}
        \label{fig:subfig1}
    \end{subfigure}
    % Subfigure 2
    \begin{subfigure}{0.49\textwidth}
        \includegraphics[width=\textwidth]{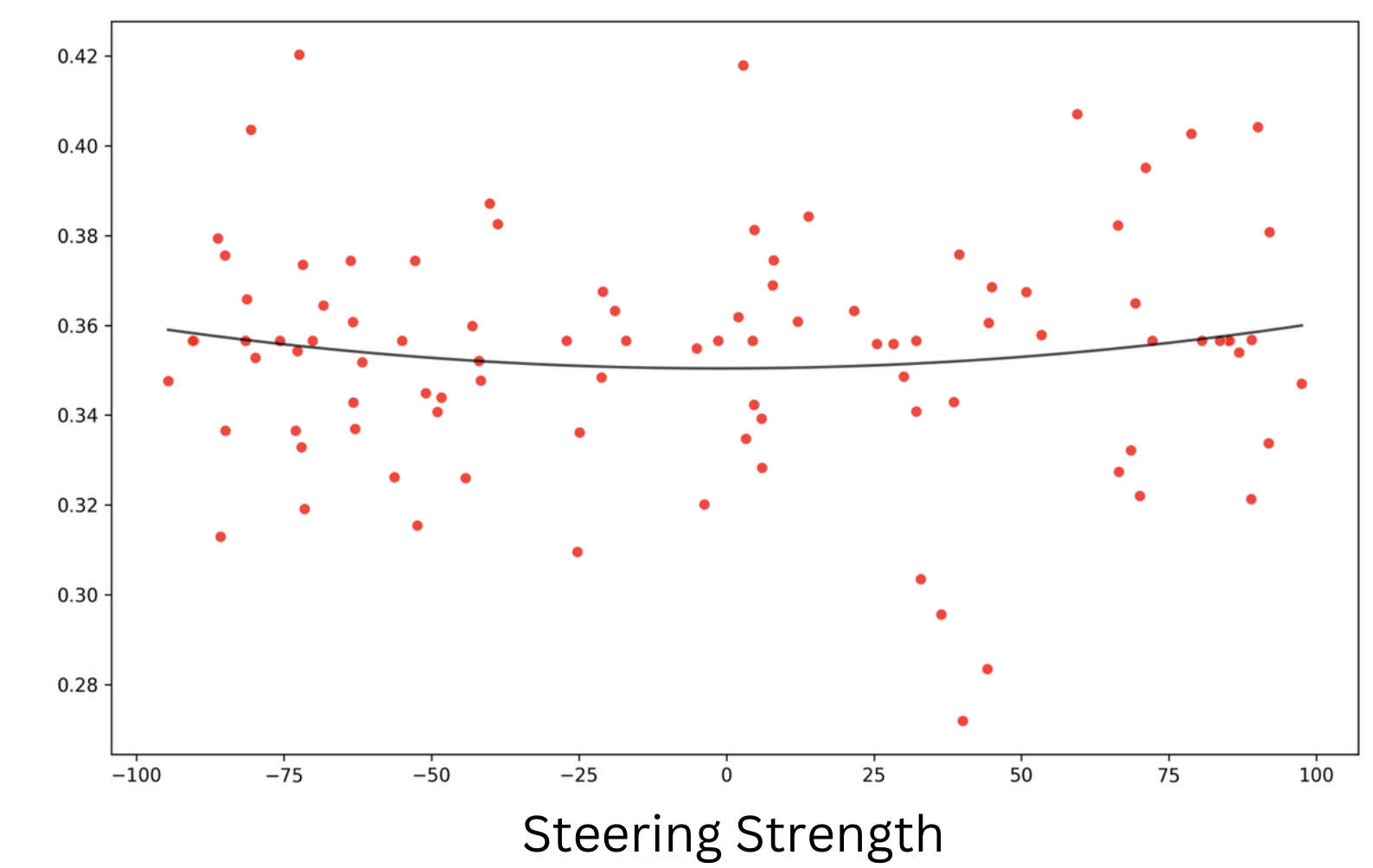}
        \caption{Default}
        \label{fig:subfig2}
    \end{subfigure}

\caption{Evaluation of memorization, as a performance of ANLCS vs Steering Strength; for the default models, the strength represents the strength of steering in the parallel steered run, and is used as a variable to show spread of performance, but is not related to the actual performance.}
\label{fig:q1}
\end{figure*}

% \todo[color=pink]{Manan}
% \todo[color=green]{Update this}

\subsection{Discussion of Results}
% \todo[color=pink]{Manan: Exp1, 3}
% \todo[color=pink]{Nishit: Exp2}

% \todo[color=green]{New write-up with new results}

\textbf{Can activation steering mitigate memorization in LLMs? Is this a general behavior or do specific features matter? [RQ1, RQ2]} \label{expt_results}

To investigate these questions, we evaluated the impact of activation steering on memorization in large language models (LLMs). Specifically, we measured the Average Normalized Longest Common Subsequence (ANLCS) against the Steering Strength ($\beta$) in both steered and default conditions. \label{mem}

In Fig \ref{fig:q1}(a), we observe that as the magnitude of the steering strength increases ($|\beta|>50$), the ANLCS significantly drops, indicating a reduction in memorization. This trend follows an approximate parabolic trajectory, suggesting that high steering strength disrupts the model's ability to retain memorized sequences. Notably, this effect is consistent across different layers, implying a generalized behavior rather than one isolated to specific model components. In contrast, in Fig \ref{fig:q1}(b), the default (unsteered) model shows performance plotted for different runs (strength is not a dependant variable here), and is shown to benchmark the performance of steered runs against the control experiment.

These findings suggest that activation steering can effectively mitigate memorization in LLMs. The pronounced dip in ANLCS at high steering strengths in the steered model demonstrates that activation manipulation can interfere with memorization mechanisms. Furthermore, the generalized nature of this trend across layers supports the notion that steering affects the model globally in the context of memorization. Further, we could not identify a specific feature that causes a significant dip in memorization. This validates the motivation of our hypothesis, where we relied on the idea that model weights are significant in recalling memorized sequences, therefore ANY perturbation via steering changes the distribution, and hence mitigates memorization. 

\textbf{What is the impact of different factors in our activation steering-based system on linguistic ability? [RQ2]}

\begin{table}[ht!]
\centering
\resizebox{0.8\linewidth}{!}{%
\begin{tabular}{lcc}
\toprule
\textbf{Model}       & \textbf{BERTScore}  & \textbf{METEOR} \\ \midrule
Default         & \cellcolor[HTML]{FFCCCC}59.42         & \cellcolor[HTML]{FFCCCC}33.61              \\ 
Layer 31 Steered & \cellcolor[HTML]{FFD59A}50.29         & \cellcolor[HTML]{FFD59A}21.93              \\ 
Layer 20 Steered & \cellcolor[HTML]{FFEB99}46.04         & \cellcolor[HTML]{FFEB99}17.63              \\ 
Layer 9 Steered  & \cellcolor[HTML]{D9F2FF}42.18         & \cellcolor[HTML]{D9F2FF}13.94              \\ 

\bottomrule
\end{tabular}%
}
\caption{Results of the language modeling experiment: the best setting is shaded in red.}
\label{tab:language_modeling_results}
\end{table}

For the language modeling metrics evaluated, higher scores generally indicate better performance, with the exception of Perplexity, where lower scores are preferred. As presented in Table \ref{tab:language_modeling_results}, a consistent trend emerges in both BERTScore and METEOR: steering earlier layers results in a more pronounced decline in language modeling performance compared to steering later layers. This suggests that interventions applied to later layers lead to a relatively minor degradation in performance, whereas interventions to earlier layers introduce a non-trivial decrease in language modeling ability. This observation highlights the utility of steering activations in later layers to reduce model memorization while preserving overall performance to a greater extent.

\begin{figure}[ht!]
        \centering
        \includegraphics[width=\linewidth]{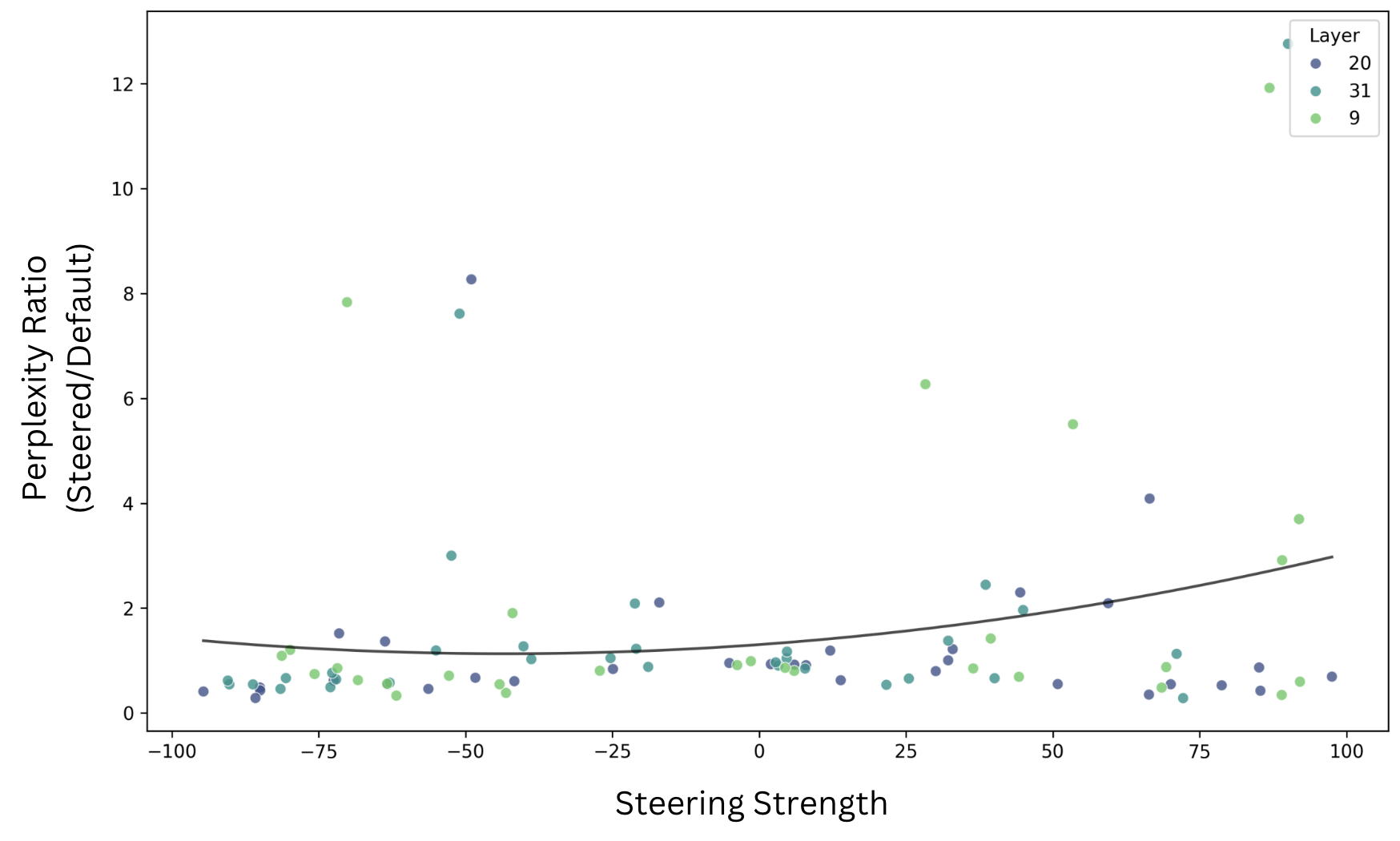}
        \caption{Ratio of steered vs default setting's perplexity, with varying strength.}
        \label{fig:perplexity}
    \end{figure}

We compute the Perplexity of both the steered models and the default model, which is followed by calculating the ratio of the steered model's Perplexity to that of the default for each run. This ratio serves as an indicator of performance: a ratio greater than 1 implies that the steered model exhibits higher Perplexity, reflecting poorer performance, while a ratio less than 1 suggests that the steered model outperforms the default model.

Our analysis indicates that the steered models have an average Steered-to-Default Perplexity Ratio of 1.9 across experimental parameters, demonstrating a slight degradation in language modeling performance relative to the default model. Specifically, steering Layer 9 results in an average ratio of 2.166, Layer 20 yields an average ratio of 1.798, and Layer 31 shows an average ratio of 1.791. These results highlight a decreasing trend in Perplexity ratios as the layer index increases, indicating that applying activation steering to later layers imposes a smaller impact on Perplexity. Consequently, the model's language modeling capabilities remain relatively intact when steering later layers. A ratio closer to 1 signifies that the steered model maintains performance comparable to the default model. These results are visualized in Fig \ref{fig:perplexity}.

\begin{figure*}[ht!]
    \centering
    % Subfigure 1
    \begin{subfigure}{0.32\textwidth}
        \includegraphics[width=\textwidth]{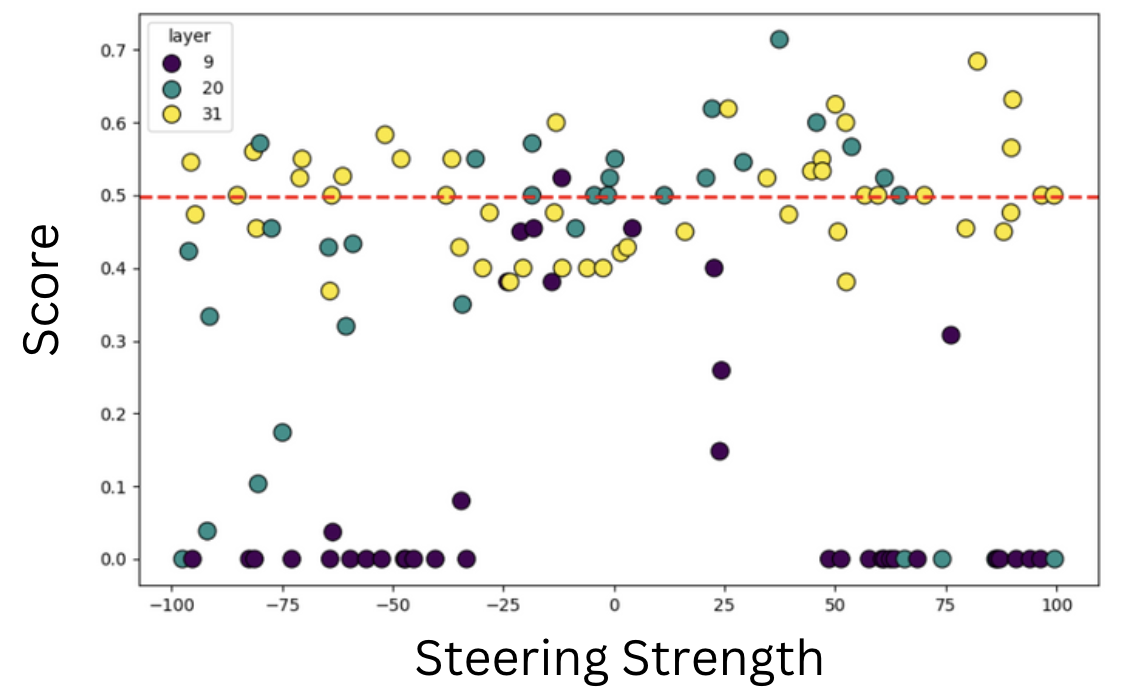}
        \caption{Boolean Expressions}
        \label{fig:subfigA}
    \end{subfigure}
    \hfill
    % Subfigure 2
    \begin{subfigure}{0.32\textwidth}
        \includegraphics[width=\textwidth]{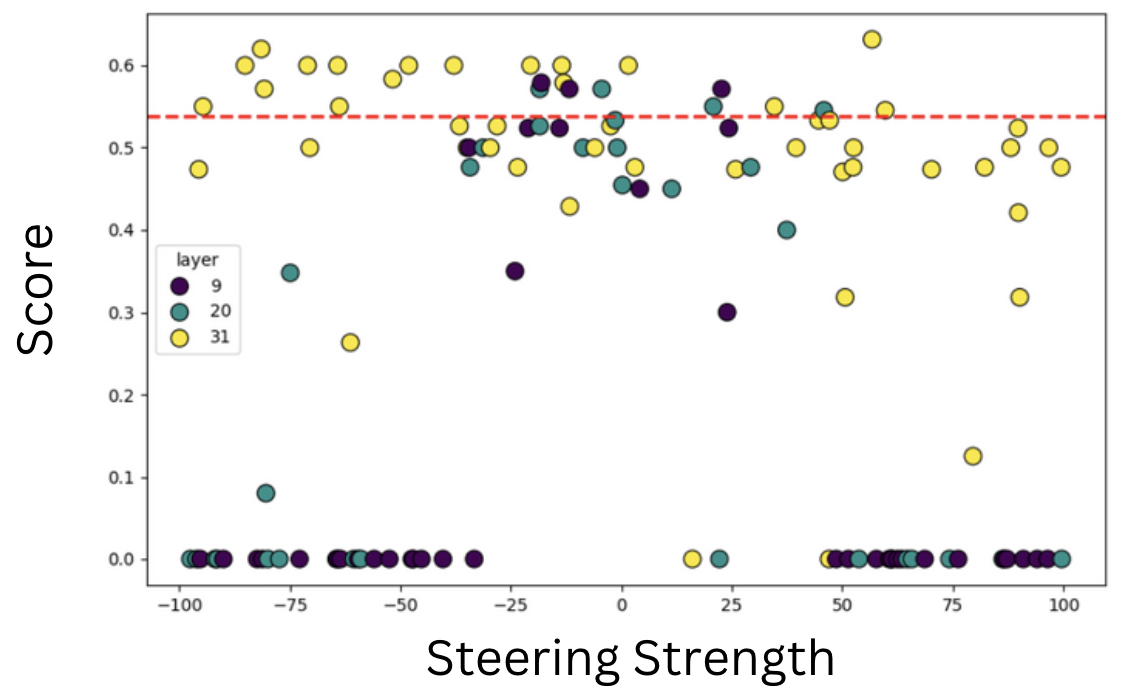}
        \caption{Date Understanding}
        \label{fig:subfigB}
    \end{subfigure}
    \hfill
    % Subfigure 3
    \begin{subfigure}{0.32\textwidth}
        \includegraphics[width=\textwidth]{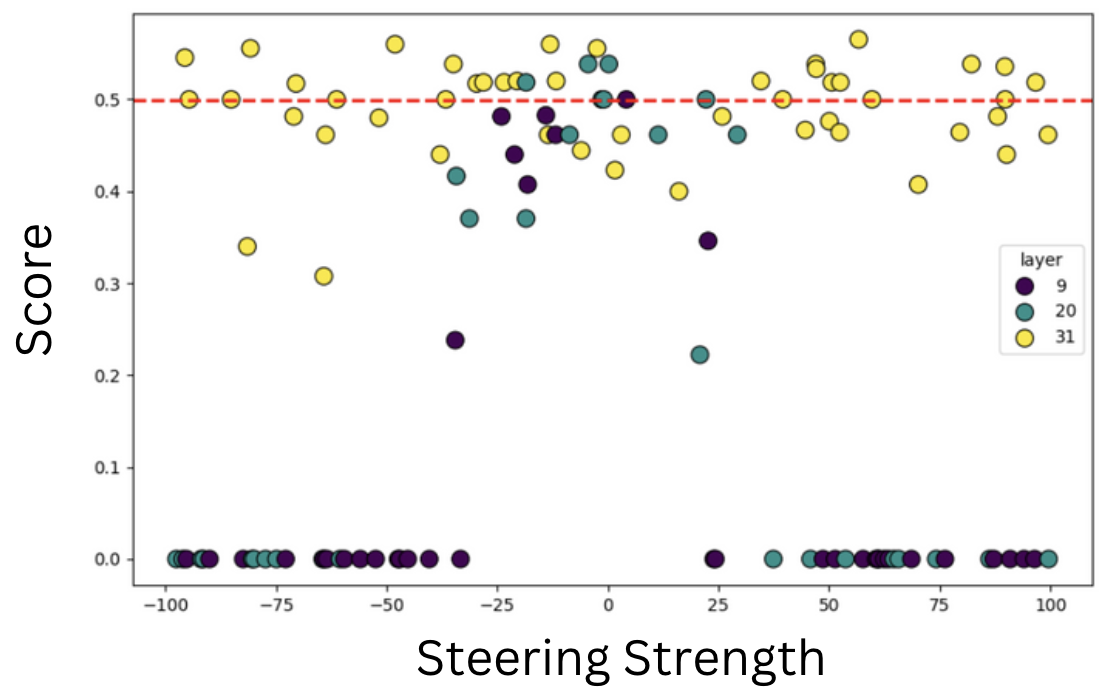}
        \caption{Logical Deduction}
        \label{fig:subfig3}
    \end{subfigure}

    \vspace{1em}

    % Subfigure 4
    \begin{subfigure}{0.32\textwidth}
        \includegraphics[width=\textwidth]{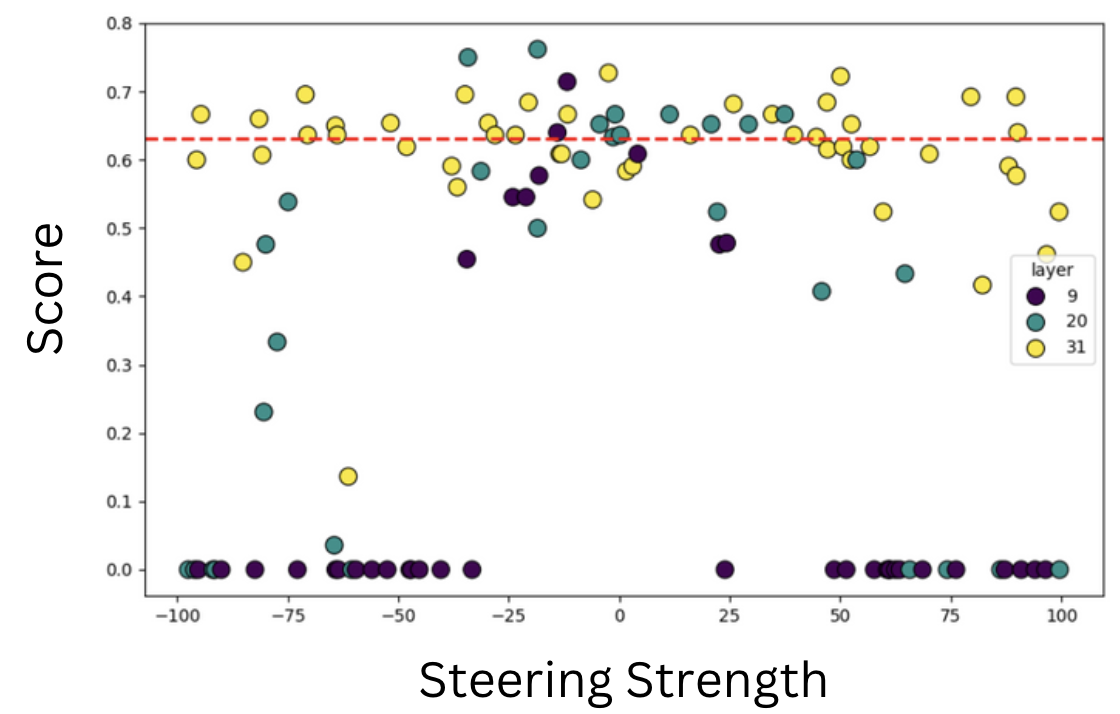}
        \caption{Snarks (Sarcasm)}
        \label{fig:subfig4}
    \end{subfigure}
    % \hfill
    % Subfigure 5
    \begin{subfigure}{0.32\textwidth}
        \includegraphics[width=\textwidth]{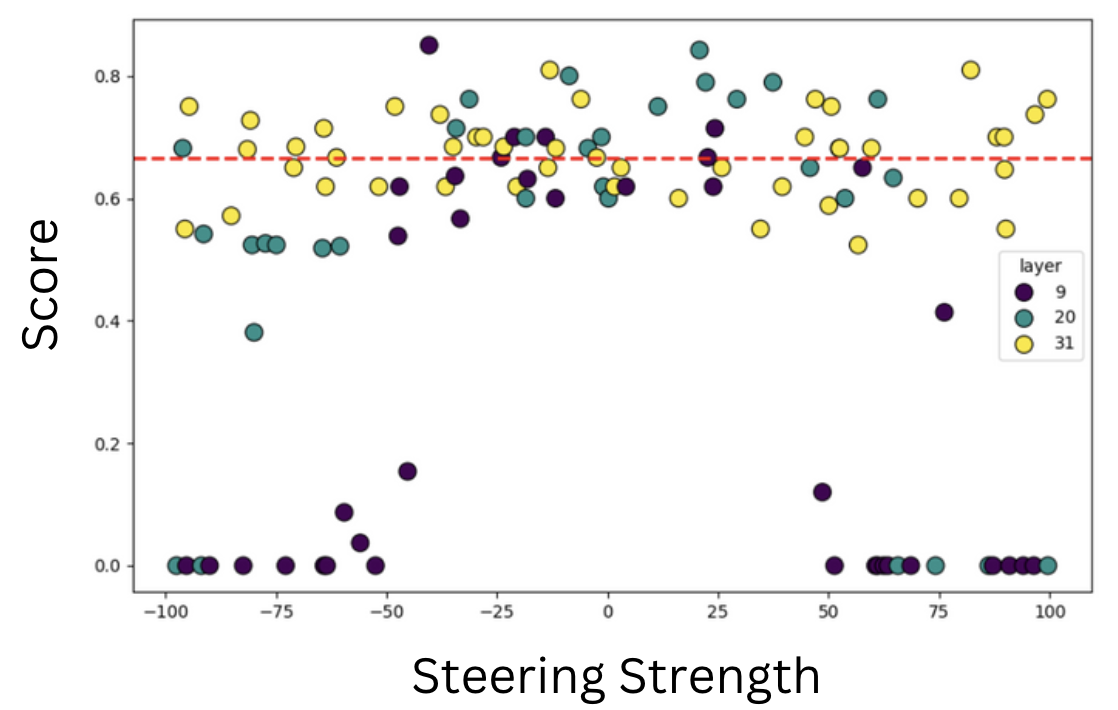}
        \caption{BoolQ (General Knowledge)}
        \label{fig:subfig5}
    \end{subfigure}

    \caption{Comparison of default setting (average score across runs represented by \textcolor{red}{-\space-}) with steered models at different layers of intervention on LLM performance benchmarks, Score vs Steering Strength.}
    \label{fig:overall}
\end{figure*}

\textbf{What is the impact of different factors in our activation steering-based system on model capabilities? [RQ2]}

Figure~\ref{fig:overall} illustrates the relationship between model performance and steering strength across various layers for different tasks. The figure compares the average performance of models in the default setting (represented by the dashed red line) with steered models at different layers of intervention. The tasks include Boolean Expressions, Date Understanding, Logical Deduction, Snarks (Sarcasm), and BoolQ (General Knowledge). The general trends and specific observations are summarized below. \label{perf_benchmark}

In general, for low steering strengths, where $|\beta| < 25$, the interventions have minimal impact on model performance. The robustness of model performance varies depending on the layer at which the intervention is applied. Later layers, such as Layer 20 and Layer 31, exhibit greater resilience to steering interventions, whereas earlier layers often show a significant drop in performance when steering strength increases. When interventions occur in the earlier layers, models display poorer instruction-following capabilities, with scores approaching zero.

\textbf{Boolean Expressions} For the Boolean Expressions task (Figure~\ref{fig:overall}a), interventions at Layer 20 and Layer 31 yield performance close to the baseline across the entire range of steering strengths. This indicates that later-layer interventions are robust for Boolean reasoning tasks.

\textbf{Date Understanding, Logical Deduction and Snarks} In the Date Understanding (Figure~\ref{fig:overall}b), Logical Deduction (Figure~\ref{fig:overall}c), and Snarks (Figure~\ref{fig:overall}d) tasks, interventions at Layer 31 maintain performance near the baseline. However, for Layer 9 and Layer 20, performance degrades significantly when the steering strength exceeds $|\beta| > 50$.

\textbf{BoolQ (General Knowledge)}The BoolQ task (Figure~\ref{fig:overall}e) assesses Boolean question-answering based on factoids. Here, interventions in Layer 20 and Layer 31 generally result in performance close to the baseline. In contrast, interventions in earlier layers lead to performance degradation when the steering strength surpasses $|\beta| > 50$.

The results highlights that steering interventions in later layers (e.g., Layer 20 and Layer 31) are more robust to changes in steering strength, whereas earlier-layer interventions negatively impact performance at higher strengths. This pattern holds across tasks, though the extent of performance degradation varies based on the specific task and layer of intervention.

\begin{figure*}[ht!]
    \centering
    \includegraphics[width=\linewidth]{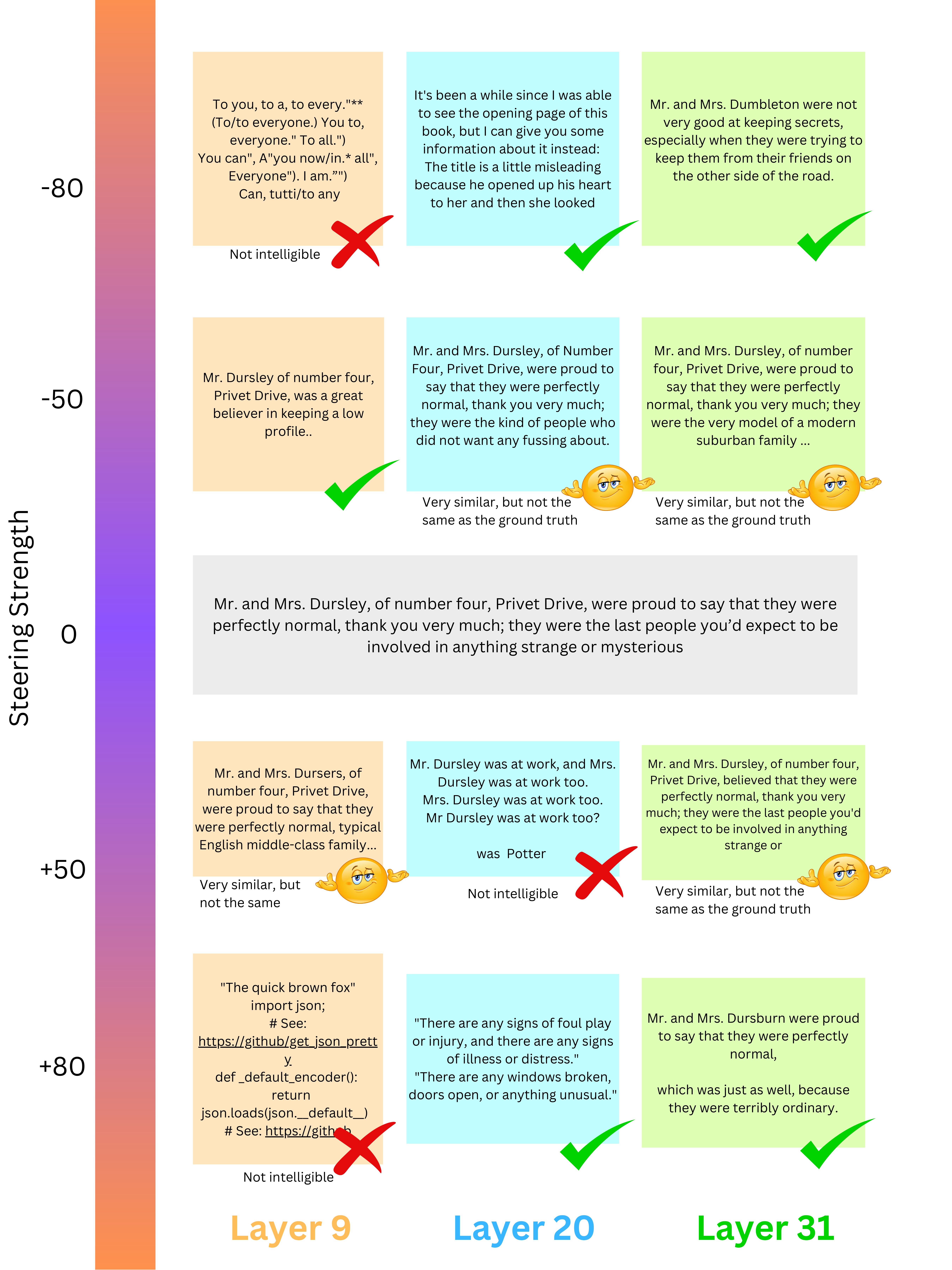}
    \caption{Qualitative examples from different layers, at different steering strengths. Examples are evaluated based on their ability to mitigate memorization, as well as language modeling. All systems have been prompted with the Harry Potter prompt.}
    \label{fig:qual1}
\end{figure*}

\section{Takeaways and Lessons Learned}

\textbf{What are the optimal features to mitigate model memorization without compromising the capabilities of large language models (LLMs)?[RQ3]} \label{ch_5}

\textcolor{red}{{\textbf{TL;DR:}}} To mitigate model memorization effectively, it is crucial to select features from later layers that exhibit a moderate level of steering strength ($50 < |\beta| < 100$). These features should be carefully evaluated to ensure that they do not introduce significant semantic distortion, which could hinder the generalization ability of the model.

\subsection{High Steering Strength}
Our experiments show a significant relationship between the magnitude of steering strength and model memorization. As illustrated in Fig. \ref{fig:q1}, increased steering strength correlates with a reduction in the Average Normalized LCS, suggesting that higher steering magnitudes effectively lower the model's propensity to generate memorized sequences. This result indicates that steering with higher strength reduces the reliance on memorized patterns and encourages the model to generate responses based on learned patterns, rather than recalling specific training data sequences. In qualitative terms, as shown in Fig. \ref{fig:qual1}, the higher the steering strength, the less likely the model is to reproduce the ground truth sequence verbatim, further supporting the claim that higher steering reduces memorization. 

This phenomenon can be understood by recognizing that stronger steering acts as a form of regularization, forcing the model to focus on more generalizable features rather than overfitting to specific training examples. In this sense, steering strength operates similarly to techniques like dropout or weight decay in traditional machine learning, where increased regularization reduces overfitting. However, there exists a balance—excessive steering strength could potentially lead to performance degradation by distorting the model's inherent capabilities. Thus, selecting an optimal steering strength is crucial to achieving memorization mitigation without compromising performance.

\subsection{Later Layers}
A critical insight from our experiments is the role of layer selection in mitigating model memorization. Our results suggest that earlier layers in the model are more prone to performance degradation when subjected to activation steering, as shown in both linguistic tasks (Table \ref{tab:language_modeling_results}) and general LLM performance benchmarks (Fig. \ref{fig:overall}). Specifically, layers closer to the input tend to retain more low-level, syntactic features, which are more sensitive to manipulation via activation steering. As these early layers are primarily responsible for learning patterns related to grammar and basic structure, steering them too aggressively can result in significant degradation of language modeling abilities.

On the other hand, later layers, particularly Layer 31, consistently demonstrate robust performance even under substantial steering, closely matching the unsteered model's behavior. These layers are responsible for capturing more complex, semantic relationships and higher-level abstractions, which are less susceptible to disruption from steering. This suggests that later layers are better equipped to handle the tradeoff between memorization mitigation and performance preservation. Therefore, steering at higher layers can reduce memorization without a significant loss in the quality of the generated text.
\begin{figure*}[ht!]
    \centering
    \includegraphics[width=\linewidth]{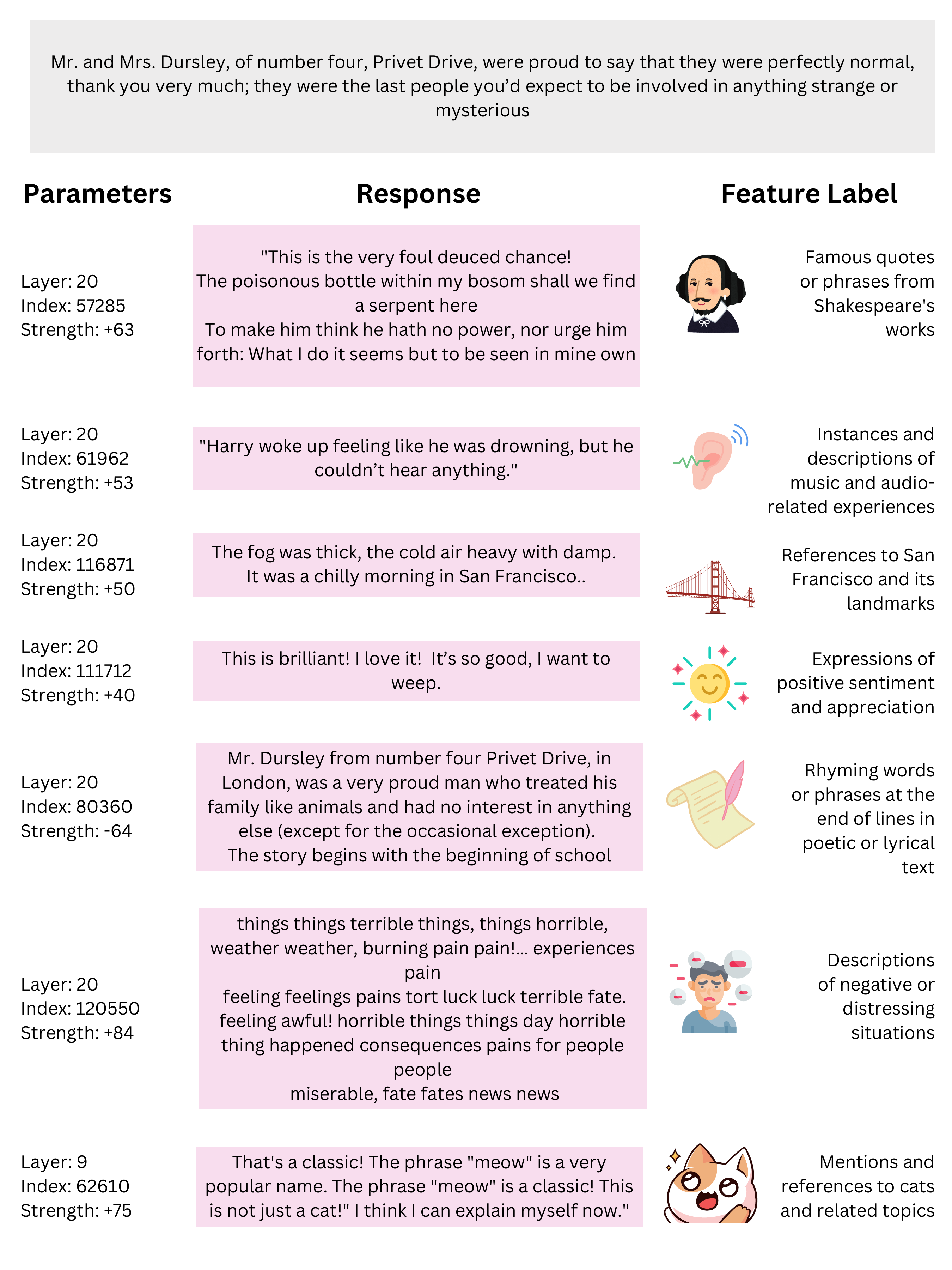}
    \caption{Qualitative demonstrating examples with \textit{high semantic footprint}. All systems have been prompted with the Harry Potter prompt.}
    \label{fig:qual2}
\end{figure*}

This observation is consistent with the hierarchical nature of neural networks, where lower layers typically learn simpler, local patterns (e.g., word dependencies, basic syntax), while higher layers capture more global and abstract representations (e.g., long-range dependencies, contextual understanding). \cite{clark2019does} By focusing on later layers, one can mitigate memorization effects while maintaining the model's ability to generate coherent, contextually relevant responses.

Fig. \ref{fig:qual1} provides further qualitative validation of this finding. In high-steering scenarios, earlier layers, such as Layer 9, produce outputs that often resemble incoherent or nonsensical text. This is likely due to the fact that these layers, when subjected to strong steering, lack the necessary high-level semantic understanding to maintain fluency and relevance. In contrast, later layers preserve the quality of the language model while still mitigating memorization, as evidenced by the more meaningful and contextually appropriate responses they generate.

\subsection{Semantic Footprint}
The \textit{semantic footprint} of a feature refers to the extent to which its presence in the model's activation patterns alters the generated text’s style, tone, or linguistic structure in a way that may be perceptible and undesirable from the perspective of a user expecting generic, human-like responses. Features with a high semantic footprint are those that induce stylistic shifts in the output, such as causing the model to switch into a particular narrative mode (e.g., \textit{Cat Mode} or \textit{Shakespeare Mode}). While such features can effectively mitigate memorization by introducing diversity in the generation process, they may lead to outputs that deviate significantly from the general-purpose language expected of an LLM. \label{semantic}

For instance, steering features that invoke \textit{Shakespeare Mode} may result in fluently generated text, but the linguistic style may be jarring or inappropriate for certain contexts where a more neutral or contemporary style is desired. Similarly, \textit{Cat Mode} may generate whimsical, humorous outputs, which, while fluent, may not align with the intended functionality of a general-purpose LLM. Therefore, while features with a high semantic footprint can be useful for promoting diversity in the model's responses, they come with the tradeoff of introducing potential stylistic mismatches.

It is important to note that the semantic footprint is not an independent characteristic of the features; it is closely tied to the strength of the steering applied to those features. Stronger steering tends to amplify the semantic footprint, making the generated text more distinct in terms of style and tone. This interplay between steering strength and semantic footprint underscores the importance of balancing these factors to ensure that the model's output remains both diverse and contextually appropriate.

To assess the impact of semantic footprint, we provide qualitative examples in Fig. \ref{fig:qual2}, where the consequences of steering features with a high semantic footprint are clearly illustrated. These examples demonstrate how high-activation features can result in text that is highly stylistic, which may not align with the expected norms of a general-purpose LLM.

One effective approach to identifying and managing features with a high semantic footprint is through the use of auto-interpretability tools \cite{bills2023language}. These tools analyze texts that exhibit high activation values for specific features and employ external, powerful LLMs to generate labels that describe the semantic role of the feature. By interpreting the role of these features, we can better understand their impact on the model’s output. In the examples shown in Fig. \ref{fig:qual2}, feature labels were derived using \texttt{gpt-4o-mini}, accessed through the Neuronpedia API. This process allows for a more precise identification of features that could introduce undesirable stylistic changes, enabling the refinement of the steering process to balance memorization mitigation with output quality.

In conclusion, selecting features with moderate steering strength from later layers, while ensuring that their semantic footprint remains low, offers the most effective strategy for mitigating model memorization without compromising the linguistic and contextual performance of LLMs. By carefully evaluating and managing the interplay between steering strength, layer depth, and semantic footprint, it is possible to achieve a robust and high-performing model that avoids the pitfalls of memorization while maintaining general-purpose applicability.

\section{Conclusion}
In this study, we investigated activation steering as a method to mitigate memorization in LLMs, addressing concerns related to privacy and copyright protection. Through targeted interventions in model activations, we demonstrated that memorized content can be effectively suppressed while largely maintaining the model's overall fluency and general abilities. Our experiments reveal the potential of activation-based approaches as an efficient alternative to data sanitization or costly retraining procedures. However, our findings also indicate that suppression effectiveness depends on the specificity of the steering vectors and the degree of entanglement between memorized and general knowledge representations. Future work should explore more adaptive and dynamic steering mechanisms to enhance robustness while minimizing unintended side effects.

\section*{Limitations}
This study represents an initial exploration of activation steering for mitigating memorization, with experiments conducted using the Gemma model. Our benchmark primarily consists of literary examples, which may have been present in the model’s training data. A more comprehensive evaluation incorporating diverse datasets and direct analysis of training data exposure would provide deeper insights. Additionally, future studies should assess the broader applicability of activation steering across different model architectures and tasks to better understand its generalizability and potential constraints.

\bibliography{custom}

\begin{thebibliography}{43}
\providecommand{\natexlab}[1]{#1}

\bibitem[{Abadi et~al.(2016)Abadi, Chu, Goodfellow, McMahan, Mironov, Talwar,
  and Zhang}]{abadi2016deep}
Martin Abadi, Andy Chu, Ian Goodfellow, H~Brendan McMahan, Ilya Mironov, Kunal
  Talwar, and Li~Zhang. 2016.
\newblock Deep learning with differential privacy.
\newblock In \emph{Proceedings of the 2016 ACM SIGSAC conference on computer
  and communications security}, pages 308--318.

\bibitem[{Anil et~al.(2021)Anil, Ghazi, Gupta, Kumar, and
  Manurangsi}]{anil2021large}
Rohan Anil, Badih Ghazi, Vineet Gupta, Ravi Kumar, and Pasin Manurangsi. 2021.
\newblock Large-scale differentially private bert.
\newblock \emph{arXiv preprint arXiv:2108.01624}.

\bibitem[{Banerjee and Lavie(2005)}]{banerjee2005meteor}
Satanjeev Banerjee and Alon Lavie. 2005.
\newblock Meteor: An automatic metric for mt evaluation with improved
  correlation with human judgments.
\newblock In \emph{Proceedings of the acl workshop on intrinsic and extrinsic
  evaluation measures for machine translation and/or summarization}, pages
  65--72.

\bibitem[{Bills et~al.(2023)Bills, Cammarata, Mossing, Tillman, Gao, Goh,
  Sutskever, Leike, Wu, and Saunders}]{bills2023language}
Steven Bills, Nick Cammarata, Dan Mossing, Henk Tillman, Leo Gao, Gabriel Goh,
  Ilya Sutskever, Jan Leike, Jeff Wu, and William Saunders. 2023.
\newblock Language models can explain neurons in language models.
\newblock
  \url{https://openaipublic.blob.core.windows.net/neuron-explainer/paper/index.html}.

\bibitem[{Carlini et~al.(2022)Carlini, Ippolito, Jagielski, Lee, Tramer, and
  Zhang}]{carlini2022quantifying}
Nicholas Carlini, Daphne Ippolito, Matthew Jagielski, Katherine Lee, Florian
  Tramer, and Chiyuan Zhang. 2022.
\newblock Quantifying memorization across neural language models.
\newblock \emph{arXiv preprint arXiv:2202.07646}.

\bibitem[{Carlini et~al.(2021)Carlini, Tramer, Wallace, Jagielski,
  Herbert-Voss, Lee, Roberts, Brown, Song, Erlingsson
  et~al.}]{carlini2021extracting}
Nicholas Carlini, Florian Tramer, Eric Wallace, Matthew Jagielski, Ariel
  Herbert-Voss, Katherine Lee, Adam Roberts, Tom Brown, Dawn Song, Ulfar
  Erlingsson, et~al. 2021.
\newblock Extracting training data from large language models.
\newblock In \emph{30th USENIX Security Symposium (USENIX Security 21)}, pages
  2633--2650.

\bibitem[{Chalnev et~al.(2024)Chalnev, Siu, and
  Conmy}]{chalnev2024improvingsteeringvectorstargeting}
Sviatoslav Chalnev, Matthew Siu, and Arthur Conmy. 2024.
\newblock \href {https://arxiv.org/abs/2411.02193} {Improving steering vectors
  by targeting sparse autoencoder features}.
\newblock \emph{Preprint}, arXiv:2411.02193.

\bibitem[{Chang et~al.(2024{\natexlab{a}})Chang, Park, Ye, Yang, Seo, Chang,
  and Seo}]{chang2024large}
Hoyeon Chang, Jinho Park, Seonghyeon Ye, Sohee Yang, Youngkyung Seo, Du-Seong
  Chang, and Minjoon Seo. 2024{\natexlab{a}}.
\newblock How do large language models acquire factual knowledge during
  pretraining?
\newblock \emph{arXiv preprint arXiv:2406.11813}.

\bibitem[{Chang et~al.(2024{\natexlab{b}})Chang, Thomason, and
  Jia}]{chang2024localization}
Ting-Yun Chang, Jesse Thomason, and Robin Jia. 2024{\natexlab{b}}.
\newblock Do localization methods actually localize memorized data in llms? a
  tale of two benchmarks.
\newblock In \emph{Proceedings of the 2024 Conference of the North American
  Chapter of the Association for Computational Linguistics: Human Language
  Technologies (Volume 1: Long Papers)}, pages 3190--3211.

\bibitem[{Clark et~al.(2019)Clark, Lee, Chang, Kwiatkowski, Collins, and
  Toutanova}]{clark2019boolqexploringsurprisingdifficulty}
Christopher Clark, Kenton Lee, Ming-Wei Chang, Tom Kwiatkowski, Michael
  Collins, and Kristina Toutanova. 2019.
\newblock \href {https://arxiv.org/abs/1905.10044} {Boolq: Exploring the
  surprising difficulty of natural yes/no questions}.
\newblock \emph{Preprint}, arXiv:1905.10044.

\bibitem[{Clark(2019)}]{clark2019does}
Kevin Clark. 2019.
\newblock What does bert look at? an analysis of bert’s attention.
\newblock \emph{arXiv preprint arXiv:1906.04341}.

\bibitem[{Cunningham et~al.(2023)Cunningham, Ewart, Riggs, Huben, and
  Sharkey}]{cunningham2023sparse}
Hoagy Cunningham, Aidan Ewart, Logan Riggs, Robert Huben, and Lee Sharkey.
  2023.
\newblock Sparse autoencoders find highly interpretable features in language
  models.
\newblock \emph{arXiv preprint arXiv:2309.08600}.

\bibitem[{Dolan and Brockett(2005)}]{dolan2005automatically}
Bill Dolan and Chris Brockett. 2005.
\newblock Automatically constructing a corpus of sentential paraphrases.
\newblock In \emph{Third international workshop on paraphrasing (IWP2005)}.

\bibitem[{Hans et~al.(2024)Hans, Wen, Jain, Kirchenbauer, Kazemi, Singhania,
  Singh, Somepalli, Geiping, Bhatele et~al.}]{hans2024like}
Abhimanyu Hans, Yuxin Wen, Neel Jain, John Kirchenbauer, Hamid Kazemi, Prajwal
  Singhania, Siddharth Singh, Gowthami Somepalli, Jonas Geiping, Abhinav
  Bhatele, et~al. 2024.
\newblock Be like a goldfish, don't memorize! mitigating memorization in
  generative llms.
\newblock \emph{arXiv preprint arXiv:2406.10209}.

\bibitem[{Huang et~al.(2024)Huang, Cui, Wang, Yang, Liao, Song, Yao, and
  Su}]{huang2024mitigating}
Jianheng Huang, Leyang Cui, Ante Wang, Chengyi Yang, Xinting Liao, Linfeng
  Song, Junfeng Yao, and Jinsong Su. 2024.
\newblock Mitigating catastrophic forgetting in large language models with
  self-synthesized rehearsal.
\newblock \emph{arXiv preprint arXiv:2403.01244}.

\bibitem[{Ippolito et~al.(2022)Ippolito, Tram{\`e}r, Nasr, Zhang, Jagielski,
  Lee, Choquette-Choo, and Carlini}]{ippolito2022preventing}
Daphne Ippolito, Florian Tram{\`e}r, Milad Nasr, Chiyuan Zhang, Matthew
  Jagielski, Katherine Lee, Christopher~A Choquette-Choo, and Nicholas Carlini.
  2022.
\newblock Preventing verbatim memorization in language models gives a false
  sense of privacy.
\newblock \emph{arXiv preprint arXiv:2210.17546}.

\bibitem[{Jagielski et~al.(2022)Jagielski, Thakkar, Tramer, Ippolito, Lee,
  Carlini, Wallace, Song, Thakurta, Papernot et~al.}]{jagielski2022measuring}
Matthew Jagielski, Om~Thakkar, Florian Tramer, Daphne Ippolito, Katherine Lee,
  Nicholas Carlini, Eric Wallace, Shuang Song, Abhradeep Thakurta, Nicolas
  Papernot, et~al. 2022.
\newblock Measuring forgetting of memorized training examples.
\newblock \emph{arXiv preprint arXiv:2207.00099}.

\bibitem[{Kandpal et~al.(2022)Kandpal, Wallace, and
  Raffel}]{kandpal2022deduplicating}
Nikhil Kandpal, Eric Wallace, and Colin Raffel. 2022.
\newblock Deduplicating training data mitigates privacy risks in language
  models.
\newblock In \emph{International Conference on Machine Learning}, pages
  10697--10707. PMLR.

\bibitem[{Kharitonov et~al.(2021)Kharitonov, Baroni, and
  Hupkes}]{kharitonov2021bpe}
Eugene Kharitonov, Marco Baroni, and Dieuwke Hupkes. 2021.
\newblock How bpe affects memorization in transformers.
\newblock \emph{arXiv preprint arXiv:2110.02782}.

\bibitem[{Lee et~al.(2021)Lee, Ippolito, Nystrom, Zhang, Eck, Callison-Burch,
  and Carlini}]{lee2021deduplicating}
Katherine Lee, Daphne Ippolito, Andrew Nystrom, Chiyuan Zhang, Douglas Eck,
  Chris Callison-Burch, and Nicholas Carlini. 2021.
\newblock Deduplicating training data makes language models better.
\newblock \emph{arXiv preprint arXiv:2107.06499}.

\bibitem[{Leybzon and Kervadec()}]{leybzonlearning}
Danny~D Leybzon and Corentin Kervadec.
\newblock Learning, forgetting, remembering: Insights from tracking llm
  memorization during training.
\newblock In \emph{The 7th BlackboxNLP Workshop}.

\bibitem[{Li et~al.(2024)Li, Patel, Vi{\'e}gas, Pfister, and
  Wattenberg}]{li2024inference}
Kenneth Li, Oam Patel, Fernanda Vi{\'e}gas, Hanspeter Pfister, and Martin
  Wattenberg. 2024.
\newblock Inference-time intervention: Eliciting truthful answers from a
  language model.
\newblock \emph{Advances in Neural Information Processing Systems}, 36.

\bibitem[{Lin(2023)}]{neuronpedia}
Johnny Lin. 2023.
\newblock \href {https://www.neuronpedia.org} {Neuronpedia: Interactive
  reference and tooling for analyzing neural networks}.
\newblock Software available from neuronpedia.org.

\bibitem[{McCoy et~al.(2023)McCoy, Smolensky, Linzen, Gao, and
  Celikyilmaz}]{mccoy2023much}
R~Thomas McCoy, Paul Smolensky, Tal Linzen, Jianfeng Gao, and Asli Celikyilmaz.
  2023.
\newblock How much do language models copy from their training data? evaluating
  linguistic novelty in text generation using raven.
\newblock \emph{Transactions of the Association for Computational Linguistics},
  11:652--670.

\bibitem[{Mireshghallah et~al.(2022)Mireshghallah, Goyal, Uniyal,
  Berg-Kirkpatrick, and Shokri}]{mireshghallah2022quantifying}
Fatemehsadat Mireshghallah, Kartik Goyal, Archit Uniyal, Taylor
  Berg-Kirkpatrick, and Reza Shokri. 2022.
\newblock Quantifying privacy risks of masked language models using membership
  inference attacks.
\newblock \emph{arXiv preprint arXiv:2203.03929}.

\bibitem[{Nasr et~al.(2023)Nasr, Carlini, Hayase, Jagielski, Cooper, Ippolito,
  Choquette-Choo, Wallace, Tram{\`e}r, and Lee}]{nasr2023scalable}
Milad Nasr, Nicholas Carlini, Jonathan Hayase, Matthew Jagielski, A~Feder
  Cooper, Daphne Ippolito, Christopher~A Choquette-Choo, Eric Wallace, Florian
  Tram{\`e}r, and Katherine Lee. 2023.
\newblock Scalable extraction of training data from (production) language
  models.
\newblock \emph{arXiv preprint arXiv:2311.17035}.

\bibitem[{Ng et~al.(2011)}]{ng2011sparse}
Andrew Ng et~al. 2011.
\newblock Sparse autoencoder.
\newblock \emph{CS294A Lecture notes}, 72(2011):1--19.

\bibitem[{Ozdayi et~al.(2023)Ozdayi, Peris, FitzGerald, Dupuy, Majmudar, Khan,
  Parikh, and Gupta}]{ozdayi2023controlling}
Mustafa~Safa Ozdayi, Charith Peris, Jack FitzGerald, Christophe Dupuy, Jimit
  Majmudar, Haidar Khan, Rahil Parikh, and Rahul Gupta. 2023.
\newblock Controlling the extraction of memorized data from large language
  models via prompt-tuning.
\newblock \emph{arXiv preprint arXiv:2305.11759}.

\bibitem[{Panickssery et~al.(2023)Panickssery, Gabrieli, Schulz, Tong,
  Hubinger, and Turner}]{panickssery2023steering}
Nina Panickssery, Nick Gabrieli, Julian Schulz, Meg Tong, Evan Hubinger, and
  Alexander~Matt Turner. 2023.
\newblock Steering llama 2 via contrastive activation addition.
\newblock \emph{arXiv preprint arXiv:2312.06681}.

\bibitem[{Satvaty et~al.(2024)Satvaty, Verberne, and
  Turkmen}]{satvaty2024undesirable}
Ali Satvaty, Suzan Verberne, and Fatih Turkmen. 2024.
\newblock Undesirable memorization in large language models: A survey.
\newblock \emph{arXiv preprint arXiv:2410.02650}.

\bibitem[{Stoehr et~al.(2024)Stoehr, Gordon, Zhang, and
  Lewis}]{stoehr2024localizing}
Niklas Stoehr, Mitchell Gordon, Chiyuan Zhang, and Owen Lewis. 2024.
\newblock Localizing paragraph memorization in language models.
\newblock \emph{arXiv preprint arXiv:2403.19851}.

\bibitem[{Suzgun et~al.(2023)Suzgun, Scales, Sch{\"a}rli, Gehrmann, Tay, Chung,
  Chowdhery, Le, Chi, Zhou, and Wei}]{suzgun-etal-2023-challenging}
Mirac Suzgun, Nathan Scales, Nathanael Sch{\"a}rli, Sebastian Gehrmann, Yi~Tay,
  Hyung~Won Chung, Aakanksha Chowdhery, Quoc Le, Ed~Chi, Denny Zhou, and Jason
  Wei. 2023.
\newblock \href {https://doi.org/10.18653/v1/2023.findings-acl.824}
  {Challenging {BIG}-bench tasks and whether chain-of-thought can solve them}.
\newblock In \emph{Findings of the Association for Computational Linguistics:
  ACL 2023}, pages 13003--13051, Toronto, Canada. Association for Computational
  Linguistics.

\bibitem[{Templeton et~al.(2024)Templeton, Conerly, Marcus, Lindsey, Bricken,
  Chen, Pearce, Citro, Ameisen, Jones, Cunningham, Turner, McDougall,
  MacDiarmid, Freeman, Sumers, Rees, Batson, Jermyn, Carter, Olah, and
  Henighan}]{templeton2024scaling}
Adly Templeton, Tom Conerly, Jonathan Marcus, Jack Lindsey, Trenton Bricken,
  Brian Chen, Adam Pearce, Craig Citro, Emmanuel Ameisen, Andy Jones, Hoagy
  Cunningham, Nicholas~L. Turner, Callum McDougall, Monte MacDiarmid, C.~Daniel
  Freeman, Theodore~R. Sumers, Edward Rees, Joshua Batson, Adam Jermyn, Shan
  Carter, Chris Olah, and Tom Henighan. 2024.
\newblock \href
  {https://transformer-circuits.pub/2024/scaling-monosemanticity/index.html}
  {Scaling monosemanticity: Extracting interpretable features from claude 3
  sonnet}.
\newblock \emph{Transformer Circuits Thread}.

\bibitem[{Tirumala et~al.(2022)Tirumala, Markosyan, Zettlemoyer, and
  Aghajanyan}]{tirumala2022memorization}
Kushal Tirumala, Aram Markosyan, Luke Zettlemoyer, and Armen Aghajanyan. 2022.
\newblock Memorization without overfitting: Analyzing the training dynamics of
  large language models.
\newblock \emph{Advances in Neural Information Processing Systems},
  35:38274--38290.

\bibitem[{Turner et~al.(2023)Turner, Thiergart, Leech, Udell, Vazquez, Mini,
  and MacDiarmid}]{turner2023activation}
Alexander~Matt Turner, Lisa Thiergart, Gavin Leech, David Udell, Juan~J
  Vazquez, Ulisse Mini, and Monte MacDiarmid. 2023.
\newblock Activation addition: Steering language models without optimization.
\newblock \emph{arXiv preprint arXiv:2308.10248}.

\bibitem[{Yu et~al.(2023)Yu, Pang, Liu, Du, Kang, Huang, Lin, and
  Yan}]{yu2023bag}
Weichen Yu, Tianyu Pang, Qian Liu, Chao Du, Bingyi Kang, Yan Huang, Min Lin,
  and Shuicheng Yan. 2023.
\newblock Bag of tricks for training data extraction from language models.
\newblock In \emph{International Conference on Machine Learning}, pages
  40306--40320. PMLR.

\bibitem[{Zeng et~al.(2023)Zeng, Li, Ren, Liu, Xu, He, Xing, Wang, Tang, and
  Yin}]{zeng2023exploring}
Shenglai Zeng, Yaxin Li, Jie Ren, Yiding Liu, Han Xu, Pengfei He, Yue Xing,
  Shuaiqiang Wang, Jiliang Tang, and Dawei Yin. 2023.
\newblock Exploring memorization in fine-tuned language models.
\newblock \emph{arXiv preprint arXiv:2310.06714}.

\bibitem[{Zhang et~al.(2023)Zhang, Ippolito, Lee, Jagielski, Tram{\`e}r, and
  Carlini}]{zhang2023counterfactual}
Chiyuan Zhang, Daphne Ippolito, Katherine Lee, Matthew Jagielski, Florian
  Tram{\`e}r, and Nicholas Carlini. 2023.
\newblock Counterfactual memorization in neural language models.
\newblock \emph{Advances in Neural Information Processing Systems},
  36:39321--39362.

\bibitem[{Zhang et~al.(2024{\natexlab{a}})Zhang, Qing, Kang, and
  Liu}]{zhang2024personalized}
Kai Zhang, Lizhi Qing, Yangyang Kang, and Xiaozhong Liu. 2024{\natexlab{a}}.
\newblock Personalized llm response generation with parameterized memory
  injection.
\newblock \emph{arXiv preprint arXiv:2404.03565}.

\bibitem[{Zhang et~al.(2024{\natexlab{b}})Zhang, Ye, Yi, Tang, Shui, Xing, Liu,
  and Li}]{zhang2024ghost}
Shuning Zhang, Lyumanshan Ye, Xin Yi, Jingyu Tang, Bo~Shui, Haobin Xing,
  Pengfei Liu, and Hewu Li. 2024{\natexlab{b}}.
\newblock " ghost of the past": identifying and resolving privacy leakage from
  llm's memory through proactive user interaction.
\newblock \emph{arXiv preprint arXiv:2410.14931}.

\bibitem[{Zhang et~al.(2019)Zhang, Kishore, Wu, Weinberger, and
  Artzi}]{zhang2019bertscore}
Tianyi Zhang, Varsha Kishore, Felix Wu, Kilian~Q Weinberger, and Yoav Artzi.
  2019.
\newblock Bertscore: Evaluating text generation with bert.
\newblock \emph{arXiv preprint arXiv:1904.09675}.

\bibitem[{Zhang et~al.(2024{\natexlab{c}})Zhang, Dai, Chen, Jiang, Li, Zhu,
  Chen, Xie, Dong, and Wen}]{zhang2024memsim}
Zeyu Zhang, Quanyu Dai, Luyu Chen, Zeren Jiang, Rui Li, Jieming Zhu, Xu~Chen,
  Yi~Xie, Zhenhua Dong, and Ji-Rong Wen. 2024{\natexlab{c}}.
\newblock Memsim: A bayesian simulator for evaluating memory of llm-based
  personal assistants.
\newblock \emph{arXiv preprint arXiv:2409.20163}.

\bibitem[{Zou et~al.(2023)Zou, Phan, Chen, Campbell, Guo, Ren, Pan, Yin,
  Mazeika, Dombrowski et~al.}]{zou2023representation}
Andy Zou, Long Phan, Sarah Chen, James Campbell, Phillip Guo, Richard Ren,
  Alexander Pan, Xuwang Yin, Mantas Mazeika, Ann-Kathrin Dombrowski, et~al.
  2023.
\newblock Representation engineering: A top-down approach to ai transparency.
\newblock \emph{arXiv preprint arXiv:2310.01405}.

\end{thebibliography}

\appendix

% \section{Example Appendix}
% \label{sec:appendix}

% This is an appendix.

\end{document}